\documentclass[10pt,twocolumn,letterpaper]{article}

\usepackage{cvpr}
\usepackage{times}
\usepackage{epsfig}
\usepackage{graphicx}
\usepackage{amsmath}
\usepackage{amssymb}

\usepackage{graphicx}
\usepackage{subcaption}
\usepackage{float}
\usepackage{placeins}
\usepackage{titling}

\usepackage[pagebackref=true,breaklinks=true,letterpaper=true,colorlinks,bookmarks=false]{hyperref}

\cvprfinalcopy % *** Uncomment this line for the final submission

\def\cvprPaperID{1887} % *** Enter the CVPR Paper ID here

\begin{document}

%%%%%%%%% TITLE
\title{InterpoNet, A brain inspired neural network for optical flow dense interpolation}

\author{Shay Zweig\\
The Gonda Multidisciplinary Brain Research Center\\
Bar Ilan University\\
{\tt\small shayzweig@gmail.com}
\and
Lior Wolf\\
The Blavatnik School of Computer Science\\
Tel Aviv University\\
{\tt\small  wolf@cs.tau.ac.il}
}
\date{}
\maketitle
%\thispagestyle{empty}

%%%%%%%%% ABSTRACT
\begin{abstract}
   Sparse-to-dense interpolation for optical flow is a fundamental phase in the pipeline of most of the leading optical flow estimation algorithms. The current state-of-the-art method for interpolation, EpicFlow, is a local average method based on an edge aware geodesic distance. We propose a new data-driven sparse-to-dense interpolation algorithm based on a fully convolutional network. We draw inspiration from the filling-in process in the visual cortex and introduce lateral dependencies between neurons and multi-layer supervision into our learning process. We also show the importance of the image contour to the learning process. Our method is robust and outperforms EpicFlow on competitive optical flow benchmarks with several underlying matching algorithms. This leads to state-of-the-art performance on the Sintel and KITTI 2012 benchmarks.       
\end{abstract}

%%%%%%%%% BODY TEXT
\section{Introduction}

The leading optical flow algorithms to date, with few exceptions, are not end-to-end deep learning. While some of them employ deep matching scores for estimating the best match in image I' for every location in image I, almost all methods employ multiple steps that do not involve learning. With the current affinity toward end-to-end deep learning solutions, the existence of large training datasets, and many concurrent contributions in the field of deep optical flow and related fields, one may wonder why this is the case.

Out of the four steps of modern optical flow pipelines: matching, filtering,  interpolation and variational refinement, we focus on the third. In this step, a sparse list of matches is transformed into dense optical flow maps. It is one of the most crucial steps and without the availability of the EpicFlow method~\cite{Revaud2015EpicFlow:Flow}, which currently dominates this step, a large number of sparse matching techniques would not have been competitive enough to gain attention. 

EpicFlow is an extremely effective  method that is based on solid computer vision foundations. However, despite using sophisticated heuristics for improved runtime, it is still rather slow and as a non-learning method, it is bounded in the performance it can deliver. 
Replacing EpicFlow by a deep learning method is harder than it initially seems. Feedforward neural networks excel in analyzing image information, but neuroscience tells us that in biological networks, lateral and top-down feedback loops are involved in solving cases where the information is missing or corrupted at random locations. 

Artificial feedback networks are slower than feedforward networks, harder to train, and have not proven themselves in the practice of computer vision. We note that feedback networks with a predefined number of feedback iterations can be unrolled into deep feedforward networks with one major caveat -- while in most feedforward networks, the supervision flows from the top down, in feedback networks, the supervision occurs at each iteration. To resolve this, we equip our network with supervision at every layer.

Inspired by neuroscience, we also suggest a loss involving lateral dependencies. Here, too, we replace the process of lateral feedback during run-time with additional supervision during training. In this way, the feedforward network learns how to mimic a network with lateral feedback loops by utilizing the training labels.

Taken together, our contributions are: (a) We propose, for the first time, to the best of our knowledge, a neural network based sparse-to-dense interpolation for optical flow. Our network performs better than the current state of the art, it is robust and can be adjusted to different matching algorithms and serve as the new default interpolation method in optical flow pipelines. (b) We introduce a new lateral dependency loss, embedding the correlations between neighbors into the learning process. (c) We define a novel architecture involving detour networks in each layer of the network. The new architecture provides a substantial increase in performance. (d) We solidify the importance of motion boundaries in learning dense interpolation for optical flow. 

%-------------------------------------------------------------------------
\section{Related Work}
\paragraph{Interpolation In The Visual Cortex.}
The visual system often receives a noisy and missing input. However, it is known to robustly denoise and fill-in the gaps in the input image. This phenomenon termed - perceptual filling-in~\cite{Komatsu2006TheFilling-in.}, %was reported to occur for features like brightness\cite{Paradiso1991BrightnessFilling-in}, color\cite{Friedman1999ColorHumans}, texture and motion\cite{Ramachandran1991PerceptualVision.} in terms of occlusions~\cite{Kellman1998ACompletion.}, illusory contours and surfaces~\cite{Pinna2001SurfaceIllusion}, in the "blind spot"\cite{Ramachandran1992BlindSpots.} and in visual scotomas~\cite{DeWeerd1995ResponsesScotoma.}. 
was reported to occur for occlusions~\cite{Kellman1998ACompletion.}, illusory contours and surfaces~\cite{Pinna2001SurfaceIllusion}, in the "blind spot"\cite{Ramachandran1992BlindSpots.} and in visual scotomas~\cite{Ramachandran1991PerceptualVision.}. Different features in the visual stimulus are filled in, including brightness\cite{Paradiso1991BrightnessFilling-in}, color\cite{Friedman1999ColorHumans}, texture and motion\cite{Ramachandran1991PerceptualVision.}.
The neurophysiological mechanism underlying perceptual filling-in is still under debate. However, many have found evidence of the existence of a neuronal filling-in mechanism~\cite{Paradiso1991BrightnessFilling-in,Poort2012TheCortex.,Wachtler2003RepresentationCortex,Huang2008V1Filling-in.,Zweig2015RepresentationHoles}. In this mechanism, neurons that are retinotopically mapped to visible or salient parts of an image (such as the edges) are activated first. This initial activation is followed by a later spread to neurons that are mapped to the missing parts, resulting in a complete representation of the image \cite{DeWeerd1995ResponsesScotoma.,Zurawel2014AV1.,Komatsu2000NeuralFilling-in}. This activation spread is mediated by both lateral connections within areas in the cortex as well as top down connections~\cite{Huang2008V1Filling-in.,Poort2012TheCortex.,Zweig2015RepresentationHoles}. It was also shown to be very sensitive to edges in the image, usually originating in edges and stops when encountered with edges~\cite{vonderHeydt2003SearchingFilling-in,Zweig2015RepresentationHoles}. Finally, neuronal filling-in was found to take place in multiple areas in the visual cortex hierarchy, from V1 and V2~\cite{Huang2008V1Filling-in.,Roe2005CorticalIllusion.b} via V4~\cite{Poort2012TheCortex.,Sasaki2004TheColor.} and in higher areas~\cite{Meng2005Filling-inBrain.,Mendola1999TheImaging.}.

We designed our interpolation network to incorporate three concepts inspired by neuronal filling-in: the interactions between neighbor neurons, multi-layer supervision and the importance of edges. Neighbor neurons' interactions can be modeled by recurrent connections within a layer, such as the model suggested by Liang and Hu~\cite{Liang2015RecurrentRecognition}. While the anatomic resemblance of such models to the cortex is appealing, in reality, they are unfolded to a feedforward network with shared weights. We, therefore, preferred to utilize the loss to force the interaction between neighbor neurons while using simpler, strictly feedforward networks, which were shown to perform extremely well for vision tasks while excelling in training time and simplicity.      
\paragraph{Interpolation For Optical Flow.}
Most current optical flow approaches are based on a four phase pipeline. The first phase matches pixels between the images in the image pair, based on nearest neighbor fields or feature matching techniques (hand engineered or learned) ~\cite{Chen2016FullGrids,HuYinlinandSongRuiandLi2016,Menze2015DiscreteFlow}. The second phase filters matches with low confidence, producing a noisy and missing flow map\cite{Bailer2016FlowEstimation}. The missing pixels usually undergo large displacements, a significant shift in appearance or are occluded in one of the images. Therefore, a third phase is needed to interpolate the missing parts and  reduce the noise. A fourth and final phase applies refinement to the interpolated dense map from Phase 3. 

The best and most used algorithm for optical flow interpolation (the third phase) is currently EpicFlow~\cite{Revaud2015EpicFlow:Flow}. EpicFlow computes the flow of each pixel using a weighted sum of the pixel's local environment. Locality is defined by a geodesic distance function based on the image edges that correspond to the motion boundaries. This edges aware approach yields good interpolation results for occluded pixels and large displacement. EpicFlow excels in interpolation. However, it is less robust to noisy matches, especially in the vicinity of large missing regions, as displayed in their Figure 8. This sensitivity to noise is increased by the fact that the noise produced by each matching algorithm displays slightly different patterns. To overcome these difficulties, a trained algorithm like ours that learns the noise patterns is more suitable. We suggest a new interpolation method based on a deep convolutional neural network. The method is applied in a feedforward manner and leads to an improvement in both accuracy and speed over the EpicFlow method.

Finally, it is noteworthy that some of the new optical flow methods do not rely on the aforementioned pipeline \cite{Sevilla-Lara2016OpticalLayers,Hur2016JointSegmentation}. One interesting example is presented by Dosovitskiy et al.~\cite{Dosovitskiy2016FlowNet:Networks} in their FlowNet model. They present an end to end convolutional neural network for optical flow that outputs a dense flow map. While their method does not reach the state of the art performance, it runs in real-time and demonstrates the power of feedforward deep learning in optical flow estimation. 

\section{Network Architecture}\label{sec:net_arch}
\begin{figure*}
\centering
% \begin{center}
\includegraphics[width=.95\linewidth]{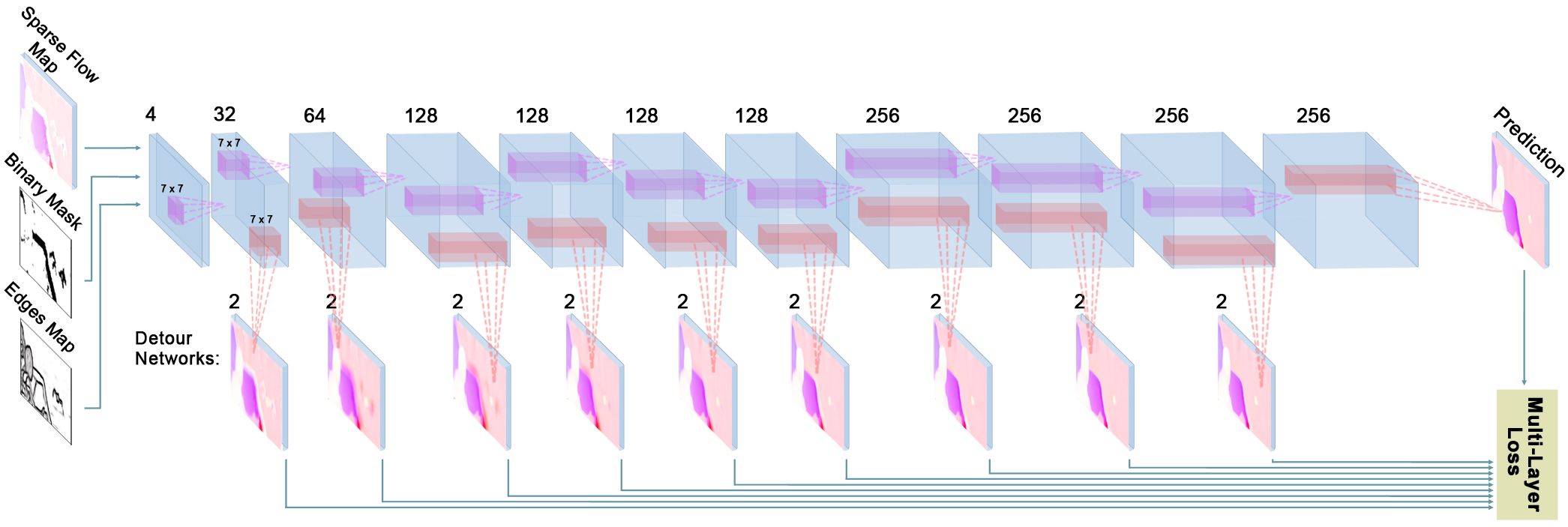}
% \end{center}
   \caption{The architecture of InterpoNet.}
\label{fig:net_arch}
\end{figure*}

The optical flow dense interpolation problem is defined in the following way: given a sparse and noisy set of matches between pixels $M=\{(p_m, p'_m)\}$, we want to approximate the dense flow field $F:I\rightarrow I'$ between a source image $I$ and a target image $I'$. To solve this problem, we use a fully convolutional network with no pooling. The main branch of the network consists of ten layers, each applying a 7x7 convolution filter followed by an Elu~\cite{clevert2015fast} non-linearity (Fig. \ref{fig:net_arch}). We use zero-padding to maintain the same image dimensions at each layer of the network.  

\subsection{Network Input}
The input to our algorithm is a set of sparse and noisy matches $M$. These matches can be produced by any third party matching algorithm. In our experiments, we used several of the leading matching algorithms: FlowFields (FF)~\cite{Bailer2016FlowEstimation},  CPM-Flow (CPM)~\cite{HuYinlinandSongRuiandLi2016}, DiscreteFlow (DF)~\cite{Menze2015DiscreteFlow}, and finally DeepMatching (DM)~\cite{Weinzaepfel2013DeepFlow:Matching}. From the matches, we produce a sparse flow map of size $h\times w\times 2$ where $h$ and $w$ are the height and width of the image pair. Each pixel is initialized with the displacement to its match in the x and y axis. Missing pixels are filled with zeros. Apart from the sparse flow map, we add two additional matrices as guiding inputs to the networks: A binary mask of the missing pixels, and the edges map (Fig. \ref{fig:net_arch}). 

We create a binary mask of all the missing pixels to indicate their position to the network (since zero can be a valid displacement value). It was shown by others~\cite{Kohler2014Mask-specificNetworks} to enhance performance in deep neural networks for inpainting. 

The last input to the network is an edges map of one of the images in the image pair for which the flow is computed. The contours of an image were shown to be a key feature in image processing in the early visual cortex~\cite{Friedman2003TheCortex.,Zurawel2014AV1.,Zweig2015RepresentationHoles,Dai2012RepresentationCortex,Poort2012TheCortex.}. EpicFlow~\cite{Revaud2015EpicFlow:Flow} already showed the benefit of the image edges as motion boundaries for optical flow estimation. In our work, we show evidence that a learning system also benefits from receiving the edges as input (see Fig.~\ref{fig:edges}). We used an off-the-shelf edges detector - the "structured edges detector" (SED)~\cite{Dollar2013StructuredDetection} - the same one used by EpicFlow. 

All of the inputs are stacked together and downsampled by 8 to form an $h/8\times w/8\times 4$ matrix. Rather than a simple stacking, we also considered different ways of introducing the edges map into the network. Among others, we have tried feeding the edges to all layers in the deep network, feeding the map to a different network and combining its output with the main branch in a deeper layer as well as constructing different networks to deal with pixels around the edges and far from the edges. However, we found that the simplest approach used here produced the best results.  
 
\subsection{The lateral dependency loss}
To optimize the network results, we used the EPE (End Point Error) loss function, which is one of the standard error measures for optical flow. It is defined as the Euclidean distance between two flow pixels:
\begin{equation}
EPE(p_1,p_2) = {||p_1-p_2||}_2
\end{equation} \label{eq:epe}
The loss for an image pair was the average EPE over pixels:
\begin{equation}
L_{epe} = \frac{1}{n} \sum_{i,j}EPE(\hat{Y}_{i,j},Y_{i,j}) 
\end{equation} \label{eq:epe_loss}\noindent
Where $\hat{Y}$ is the network prediction, $Y$ is the ground truth flow map and n is the number of pixels in the flow map.

This standard loss by itself does not yield good enough results (see Sec.~\ref{sec:exp_loss_var}). We, therefore, resort to cortical neuronal filling-in processes in our search for better losses. %We therefore used two new components in the loss, both inspired by the neuronal filling-in process in the visual cortex. 

%\subsection{Lateral dependency loss}
Neuronal filling-in is characterized by spatial spread of activation. There is evidence that the activation spread is mediated by both lateral and top-down connections. To imitate the lateral dependency between neighbors in the network, we define a new lateral dependency loss. This loss pushes the distance between neighboring pixels to be similar to the distance in the ground truth flow. It is defined in the following way:
\begin{align}
L_{ld} = 
\frac{1}{n}\sum_{i,j} &|EPE(\hat{Y}_{i,j}, \hat{Y}_{i-1,j})-EPE(Y_{i,j}, Y_{i-1,j})| + \nonumber\\ &|EPE(\hat{Y}_{i,j}, \hat{Y}_{i,j-1})-EPE(Y_{i,j}, Y_{i,j-1})| \nonumber
\end{align} \label{eq:lateral_lass}
The proposed loss term directly includes the local spatial dependencies within the training process, similar to what happens in the early stages of the visual cortex~\cite{Alonso2002NeuralCortex,Huang2008V1Filling-in.}.% This additional term added to the loss boosted the network performance as shown in section \ref{sec:exp_loss_var}.  

\subsection{Multi-layer loss using detour networks}
Top-down connections are tricky to implement in artificial neural networks. % and limited success has been shown in such architectures.
We, therefore, use the loss function, which is the main feedback to the network, to imitate top-down connections. Also inspired by the evidence that neuronal filling-in takes place in many layers in the visual system hierarchy~\cite{Poort2012TheCortex.,Meng2005Filling-inBrain.,Mendola1999TheImaging.}, we used detour networks connecting each and every layer directly to the loss function. 

During training, the loss function served as top down information pushing each layer to perform interpolation in the best possible manner. The detour networks were kept simple: aside from the main branch of the network, each of the layer's activations was transformed into a two channels flow map using a single convolution layer with linear activations (no nonlinearity, see Fig. \ref{fig:net_arch}). Each of the flow maps produced by the detour networks was compared to the ground truth flow map using the EPE and LD losses. The final network loss was the weighted sum of all the losses:
\begin{equation}
	L_{net}=\sum_{l}w^l (L_{epe}^l + L_{ld}^l)
\end{equation}
Where $w^l$, $L_{epe}^l$ and $L_{ld}^l$ are the weight, EPE loss and LD loss of layer l. We found that weights of 0.5 for each of the middle layers and 1 for the last yielded the best results. For inference, we use only the last detour layer output - the one connected to the last layer of the network's main branch.   
 
Our approach has some similarities to the one used in the inception model introduced by Szegedi et al.~\cite{Szegedy2015GoingConvolutions}, which employs auxiliary networks with independent losses during training. They found it to provide regularization and combat the vanishing gradients problem. However, in their network, the first auxiliary network was added in the tenth layer. We found that adding a detour network for each layer gave the best results. Szegedi et al.'s auxiliary networks were also built of several layers and performed some computation within them. We found that the simplest linear convolution was the best architecture. Additional layers or non-linearities did not improve the performance of the network.   

\begin{figure}
\centering
% \begin{center}
\includegraphics[width=.95\linewidth]{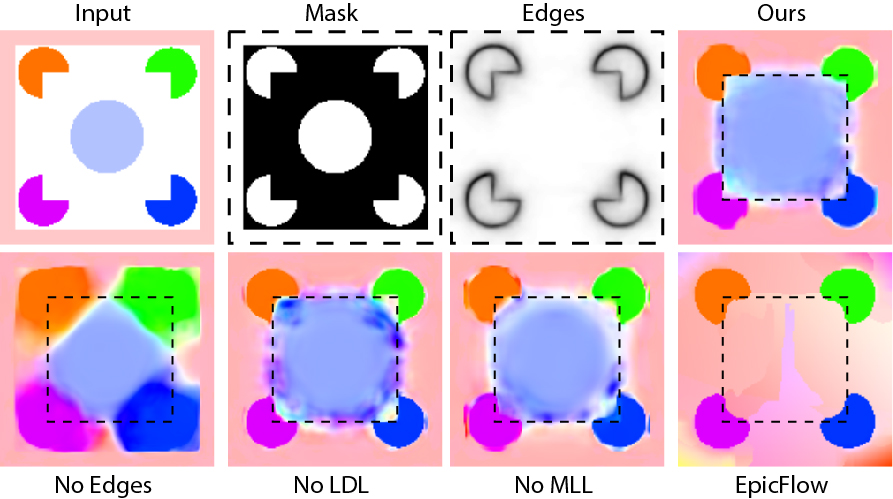}
% \end{center}
   \caption{The network prediction for the Kanizsa illusion.}
\label{fig:kanizsa}
\end{figure}

Taken together, our network was equipped with mechanisms with which it could imitate interpolation in the visual cortex. Interestingly, not only did it learn to perform interpolation of regular motion, it also performed strikingly similar to the visual cortex, when presented with an illusion. Figure \ref{fig:kanizsa} shows the interpolation applied by different variants of our network and EpicFlow on a given Kanizsa like motion pattern. The network never saw such a pattern in training time. When masking parts of the image, our network interpolates the motion pattern from the background and the interior. The propagation from the background stops in the borders of the imaginary square contour (marked by a dashed line), much like our visual perception. Importantly, only the real edges, not those of the imaginary contour, were fed to the network. Other networks that were not equipped with all the tools we presented as well as EpicFlow, performed different levels of a simpler interpolation.% with only the real edges taken into account. This illustrates the strength of a learning system compared to an engineered one.   

\subsection{Post-processing}
Our fully convolution with zero padding and no pooling network produces an output in the same size of the input. We, therefore, upsample the output by a factor of 8 using bi-linear interpolation. %approach produced an output which was similar in size to the input. We, therefore, needed to upsample the output by 8 to achieve the original image dimensions. We used bi-linear interpolation, that, despite a mild blurring effect, had an insignificant effect on our results. Other interpolations such as bi-cubic or even embedding the interpolation in the network could slightly improve the results at the cost of longer training and inference times. 
Like others before us~\cite{Dosovitskiy2016FlowNet:Networks}, we found that using the variational energy minimization post-processing used in EpicFlow~\cite{Revaud2015EpicFlow:Flow} slightly improved our final prediction (0.25px. gain in mean EPE). We employ the same parameters as EpicFlow, as appears in their Section 4. 

%Our fully convolution with zero padding and no pooling approach produced an output which was similar in size to the input. We, therefore, needed to upsample the output by 8 to achieve the original image dimensions. We used bi-linear interpolation, that, despite a mild blurring effect, had an insignificant effect on our results. Other interpolations such as bi-cubic or even embedding the interpolation in the network could slightly improve the results at the cost of longer training and inference times. Like others before us~\cite{Dosovitskiy2016FlowNet:Networks}, we found that using the variational energy minimization post-processing used in EpicFlow~\cite{Revaud2015EpicFlow:Flow} slightly improved our final prediction (~0.25 gain in mean EPE). For this phase, we use the same parameters as EpicFlow, as appear in their section 4. 

\section{Experiments}
We report the results of our network on the Sintel~\cite{sintel}, KITTI 2012~\cite{kitti2012} and KITTI 2015~\cite{Menze2015ObjectVehicles} datasets. We also show the effectiveness of different features in the network: the lateral dependency loss, the multi-layer loss and the edges input.    
\subsection{Training details}
\noindent\textbf{Preprocessing.} As described in Section \ref{sec:net_arch}, the network receives a four channel input composed of two sparse flow channels given as the output of a matching algorithm, a binary mask and the edges map. To reduce training time, we downsample all the inputs by 8 (some matching algorithms output a downsampled version by default~\cite{Weinzaepfel2013DeepFlow:Matching,Menze2015DiscreteFlow,HuYinlinandSongRuiandLi2016}). To reduce the number of missing pixels in training time, we apply bi-directional averaging (see supplementary).% materials for the details regarding downsampling and bi-directional averaging)

We apply flipping as our only data augmentation method. Other transformations such as scaling, shearing, rotating and zooming did not improve the network performance, probably due to the interpolations that accompany them and drastically change the flow map.

\noindent\textbf{Datasets.} We evaluated our network on the three main optical flow benchmarks:
MPI Sintel \cite{sintel} is a collection of several scenes taken from a graphical animation movie. Each scene consists of several consecutive frames for which a dense ground truth flow map is given (a total of 1041 training image pairs). The scenes are diverse and include battle scenes with challenging large displacements.
KITTI 2012 \cite{kitti2012} is composed of real world images taken from a moving vehicle (194 training images). And KITTI 2015, \cite{Menze2015ObjectVehicles} is similar to the KITTI 2012 dataset but with more challenging scenes (200 training images). 

Since convolutional networks demand a large set of training data, we use the same approach used by Dosovitskiy et al.~\cite{Dosovitskiy2016FlowNet:Networks}. For initial pre-training, we use the Flying Chairs dataset that they introduced. This is a relatively large synthetic dataset (22,875 image pairs) composed of chair objects flying over different backgrounds. We train on all the dataset and use a sub-sample of the Sintel dataset as validation. Due to time constraints, we could not apply all of the matching algorithms to the big flying chairs dataset. We, therefore, used only FlowFields~\cite{Bailer2016FlowEstimation} for this initial training on Flying Chairs. Additional fine tuning was applied using the training sets of specific benchmarks and for the specific matching algorithm (see supplementary). % We split each of the datasets into training and validation sets. The validation sets for both KITTI2012 and KITTI2015 datasets were the last 20\% of the pairs in each. For the Sintel dataset, due to the temporal dependencies within scenes which are a pitfall for over-fitting, we define 4 whole scenes including 167 image pairs as a validation set rather than a random sample. We use the same validation set in the pre-training and Sintel fine tuning phases. 
In all presented experiments to follow, we pre-train the networks on Flying Chairs and fine tune on Sintel using the FlowFields matching algorithm - unless stated otherwise. All the analysis, results and visualizations are done without the variational post-processing, except for the benchmark results.

\noindent\textbf{Optimization.} We use Adam~\cite{Kingma2014Adam:Optimization} with standard parameters ($\beta_1=0.9$, $\beta_2=0.999$).  A learning rate of $5\times 10^{-5}$ for the pre-training and $5\times 10^{-6}$ for the fine tuning is used.  % as the optimization method with standard parameters: $\beta_1=0.9$, $\beta_2=0.999$. Early stopping served as our only regularization method. The number of steps before performing the stop was 5000,1000 and 400 for training on the flying chairs, Sintel and KITTI datasets respectively. We use 4 rounds of early stopping in which we divide the learning rate by two starting with a learning rate of $5e^{-5}$ for the pre-training and $5e^{-6}$ for the fine tuning. After 4 rounds, we choose the weights that yielded the best performance on the validation set throughout the training. 

\subsection{Comparison of loss variants}
\label{sec:exp_loss_var}

% \begin{table}
%   \begin{minipage}[c]{0.67\linewidth}
%     \includegraphics[width=\linewidth]{LayersC.jpg}
%   \end{minipage}\hfill
%   \begin{minipage}[c]{0.3\linewidth}
%     \caption{
%        asdfsdfasdfaer gswert 
%     } \label{fig:03-03}
%   \end{minipage}
% \end{table}

\begin{table} 
%\begin{center}
\begin{minipage}[c]{0.57\linewidth}
\begin{tabular}{|l|c|}
\hline
Loss & EPE \\
\hline
EPE only& 6.104\\
EPE + LDL & 5.833\\
EPE + MLL & 5.656\\
\textbf{EPE + LDL + MLL}& \textbf{5.470} \\
\hline
\end{tabular}
\end{minipage}
\begin{minipage}[c]{0.4\linewidth}
%\end{center}
\caption{Comparing losses for the Sintel 'final' pass validation set, trained on the output of FlowFields.}
\label{tab:loss}
\end{minipage}
\end{table}

% \begin{figure*}
% \begin{center}
% \includegraphics[width=0.9\linewidth]{Layers.jpg}
% \end{center}
%    \caption{The contribution of different losses. \textbf{A}. The progression of the prediction process throughout the different layers in the network, as shown by the detour networks output. The second and third columns are the EPE and LD loss maps respectively. The last three rows are the final predictions of networks with different losses, they are presented after upsampeling which contributes to the decrease in LDL \textbf{B.} Mean EPE over different pixel groups in Sintel validation set as a function of the different layers. \textbf{C.} Comparison of the network performance with and without the LD loss. Left column is the ground truth, center and right columns are the predictions with and without LDL respectively.} 
% \label{fig:loss}
% \end{figure*}

\begin{figure}
\centering
	\begin{subfigure}{1\linewidth}
        \centering
        \includegraphics[width=\linewidth]{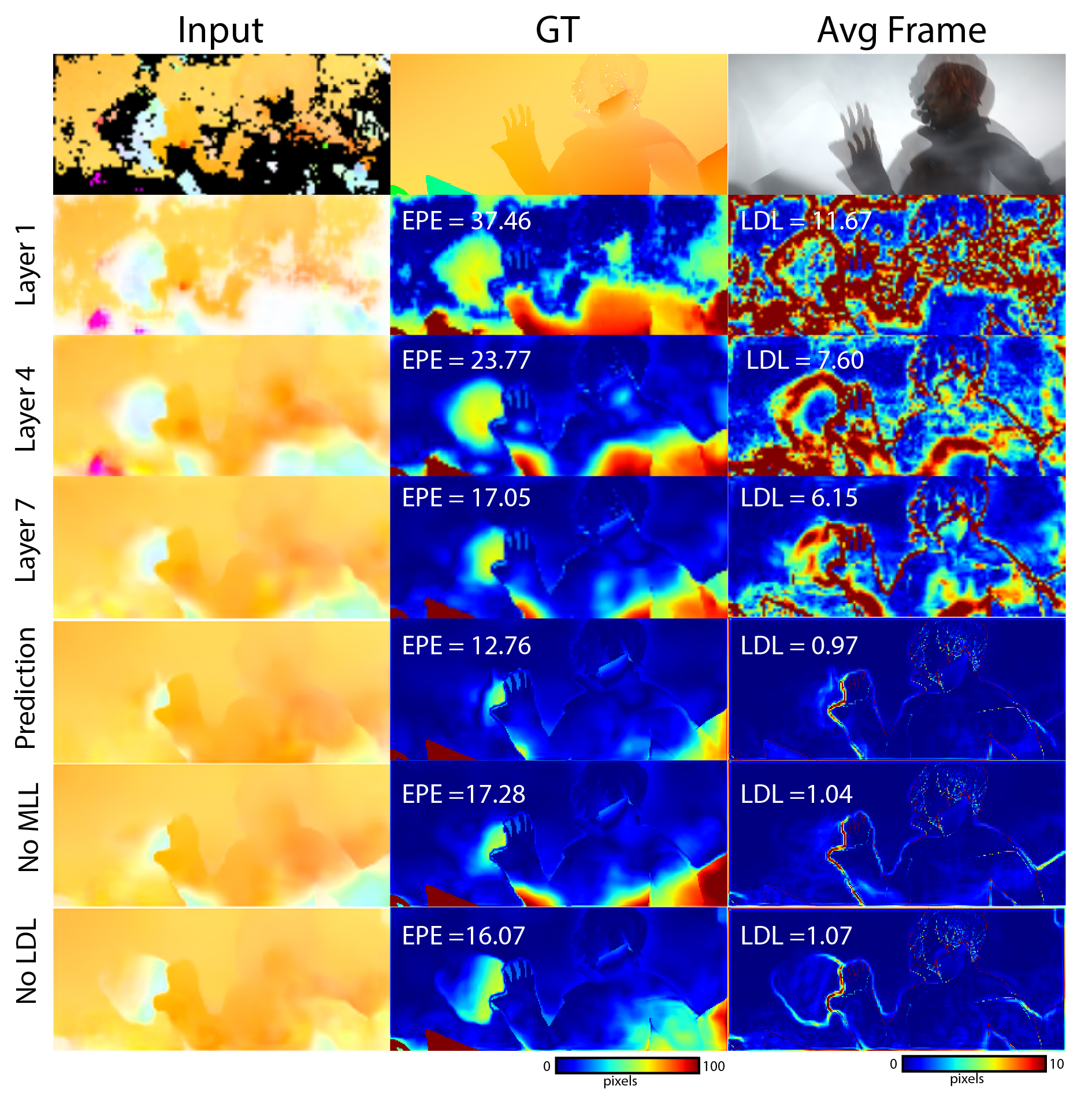}
        \caption{The progression of the prediction process throughout the different layers in the network, as shown by the detour networks outputs. Starting from the second row, the second and third columns are the EPE and LD loss maps respectively. The last three rows are the final predictions of networks with different losses. They are presented after upsampling which contributes to the decrease in LDL. Missing pixels in the input are marked in black.}\label{fig:loss_a}
    \end{subfigure} 
    \begin{subfigure}{1\linewidth}
        \centering
        \includegraphics[width=\linewidth]{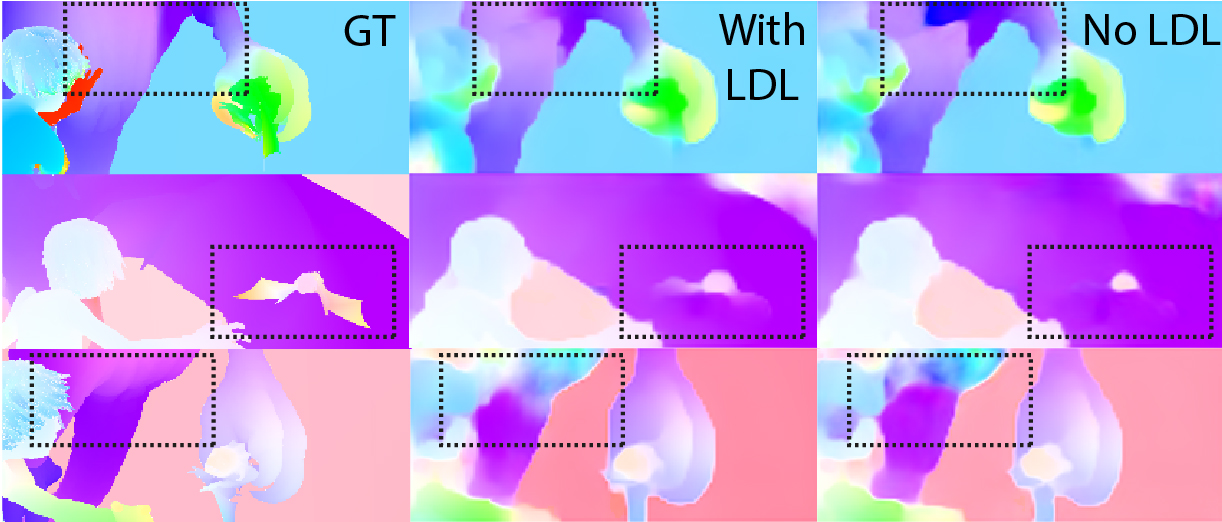}
        \caption{Comparison of the network performance with and without the LD loss. Left column is the ground truth, center and right columns are the predictions with and without LDL respectively.}\label{fig:loss_c}
    \end{subfigure} %
	\begin{subfigure}{1\linewidth}
    	\begin{minipage}{0.5\linewidth}
    	
          \centering
          \includegraphics[width=\linewidth]{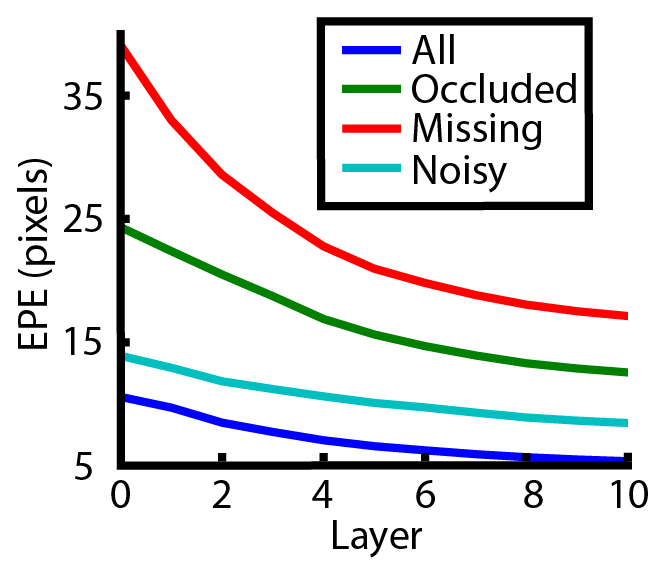}
        \end{minipage}
        \begin{minipage}{0.48\linewidth}
        	\caption{Mean EPE over different pixel groups in the Sintel validation set as a function of the different layers. Pixel groups: Noisy : with $EPE>3$ in the input matches; Occluded: appear only in one of the image pair; Missing: missing in the input matches but not occluded.}\label{fig:loss_b}
    	\end{minipage}
    \end{subfigure} %
    
	\caption{The contribution of different losses.}\label{fig:loss}
\end{figure}

To ensure the efficiency of the different losses and the new architecture we introduce, we trained several variants of our network - with only the EPE loss, with the EPE + LD loss and with the EPE + ML loss. As Table \ref{tab:loss} shows, each of our introduced losses yields a performance boost. Figure \ref{fig:loss_a} shows the output of the different detour networks in different layers as well as the error maps for the two losses we used - EPE and LD loss, for an example image in our Sintel validation set. Notice how both the EPE and LD loss improves as the network deepens - this is consistent over all of the images in the validation set (Fig. \ref{fig:loss_b}). At the first layers of the network, it seems that it is focused on performing a simple interpolation to mainly fill the missing parts. This initial interpolation is less aware of the motion boundaries. As the network deepens, it mainly polishes the details and reduces noise according to the segmentation introduced by the edges (for example. the green patches in Fig. \ref{fig:loss_a} left column). The prediction of a network trained without multi-layer loss is noisier (6th row in Fig. \ref{fig:loss_a}, lower right corner). It seems that the added supervision in all the layers helps to extinguish errors in the interpolation and adjust it according to the motion boundaries.

The LD loss is introduced to enforce a certain dependency between neighboring pixels. Much like in neuronal filling-in, lateral dependency plays a role in propagation, especially in terms of uncertainty. For example, in Figure \ref{fig:loss_a}, there is a big missing part in the center-left with some false matches (light green in the flow map) to its right. The network can choose to either propagate the background or the false matches. To avoid the wrong local dependencies, the network with LD loss uses the background to fill most of the area, almost extinguishing the false matches (notice the shrinking "bubble" in the LDL maps in Fig. \ref{fig:loss_a} rows 2-5). The network without the LD loss does not use this information and leaves a high contrast where it should not appear (Fig. \ref{fig:loss_a} last row). LD loss mostly enforces smoothness in the outcome (Fig. \ref{fig:loss_c} first row), but it also encourages high contrasts where they should appear, as shown in the example in Figure \ref{fig:loss_c} (second row) for the wings of the small dragon. In rear cases, the LD loss combined with a poor input could decrease performance like in the third row of Figure \ref{fig:loss_c} where the smooth transition introduced by the LD loss decreased the performance. Overall, the LD loss improves the EPE in 60\% of predictions in our validation set, and 80\% out of the noisy examples in the set (those with over one percent of noisy and missing pixels).

\subsection{The importance of the edges}
\label{sec:exp_edges}
\begin{table} 
% \begin{center}
\centering
\begin{tabular}{|l|c|}
\hline
Input & EPE \\
\hline
Sparse map + mask& 6.240\\
\textbf{Sparse map + mask + edges map}& \textbf{5.470} \\
\hline
\end{tabular}
% \end{center}
\caption{Comparison of the network results with and without the edges as input. The reported EPE is for the Sintel 'final' pass validation set. The network is trained on the FlowFields algorithm output.}
\label{tab:edges}
\end{table}
% \begin{figure}[t]
% \begin{center}
% \includegraphics[width=1\linewidth]{Edges.jpg}
% \end{center}
%    \caption{The contribution of the edges input to the network. \textbf{A}. A comparison between the predictions of the network with and without the edges input on two examples from the Sintel validation set. Edges are marked with black lines. \textbf{B.} Mean improvement index over all pixel validation set for missing (blue) and non missing (green) pixels.} 
% \label{fig:edges}
% \end{figure}

\begin{figure*}[t]
\centering
	\begin{subfigure}{0.7\linewidth}
        \centering
        \includegraphics[width=\linewidth]{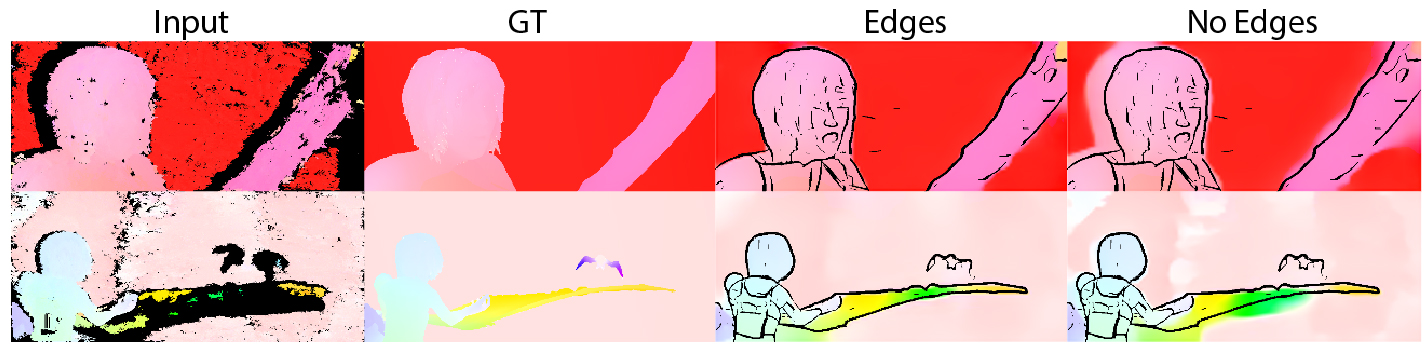}
        \caption{}\label{fig:edges_a}
    \end{subfigure} %
    \begin{subfigure}{0.25\linewidth}
        \centering
        \includegraphics[width=\linewidth]{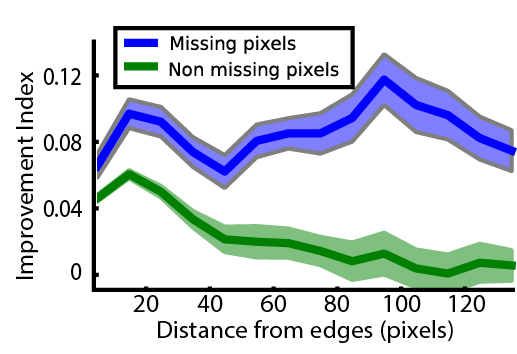}
        \caption{}\label{fig:edges_b}
    \end{subfigure} %
    \caption{The contribution of the edges input to the network. (a) The predictions of the networks with and without the edges input. Edges are marked with black lines. (b) Mean improvement index over missing (blue) and non-missing (green) pixels. Shaded areas marks $\pm$ SEM over image pairs in the Sintel validation set.}\label{fig:edges}
\end{figure*}

To validate the importance of the edges as an input to the network, we perform an experiment in which the edges are not fed into the network (in both train and test time). Table \ref{tab:edges} shows the significant impact the edges input has on the performance of the network. Much like neuronal filling-in in the visual cortex, our network uses the edges as a boundary for local spread in missing or occluded areas. Figure \ref{fig:edges_a} shows a comparison between the prediction of the two networks (with and without the edges input) for two examples from the Sintel validation set. Notice the spread into the missing pixels, while the network without the edges input performs what seems like a simple interpolation from all of the surroundings, the network with the edges input uses this information and stops the spread at the edges. 

To quantify the effect of the edges on the missing and occluded pixels, we define an improvement index (II):
\begin{equation}
	II_{p} = \dfrac{EPE_{p-noedges}-EPE_{p-edges}}{EPE_{p-noedges}+EPE_{p-edges}}  
\end{equation}
where $EPE_{p-noedges}$ and $EPE_{p-edges}$ are the EPE in pixel p between the prediction of the edges network and non-edges network respectively. Positive values of this index indicate improvement as a result of the edges input, while negative values indicate a decrease in performance. Mean II over occluded and missing pixels is significantly higher than the mean II over the non-missing pixels in the Sintel validation set ($Mean\pm SEM$ II difference = $0.0235\pm5.83\times 10^{-3} $ ;paired t-test $p<1\times 10^{-4}$ ;n=167). Interestingly, as demonstrated in Figure \ref{fig:edges_b},  the contribution of the edges input to the performance in the occluded and missing areas is not influenced by the distance from the edges. This is expected, since the decision about the spread is dependent on the segmentation by the edges and, therefore, even far away from the edges the effect is considerable (see top right corner in our prediction in the first row of Fig. \ref{fig:epic} as an example). For non-missing pixels, however, the performance gains decrease almost monotonically with the distance from the edges (green line in Fig. \ref{fig:edges_b}). These pixels are processed differently in the network, since they have initial values. They are more affected by their immediate surroundings. Therefore, a nearby edge can improve their prediction but less so far from edges. In all distances, the II values are significantly higher for the missing and occluded pixels (Wilcoxon signed rank test $p< 0.05$). 
\subsection{Fine tuning}
\label{sec:exp_finetuning}
\begin{table} 
% \begin{center}
\centering
\begin{tabular}{|l|l|c|c|c|}
\hline
\multicolumn{1}{|c|}{pre-training} & \multicolumn{1}{|c|}{Evaluated}  &  \multicolumn{3}{|c|}{Fine tuned on}\\\cline{3-5}
  & \multicolumn{1}{|c|}{on} &   None&  FF & Self \\
\hline
fc FF & Sintel FF & 5.802 & 5.470 & - \\
fc FF & Sintel CPM & 6.165 & 5.782 & 5.851 \\
fc FF & Sintel DM & 6.498 & 6.075 & 5.971 \\
fc FF & Sintel DF & 6.543 &  6.35 & 6.142 \\
\hline
fc DM & Sintel DM & 6.665 & - & 5.934 \\
\hline
\end{tabular}
% \end{center}
\caption{The network performance (EPE) without fine tuning, with FlowFields fine tuning and with fine tuning for the specific matching algorithm used for evaluation. EPE is reported for the Sintel 'final' pass validation set. Notations: fc - Flying Chairs, FF -FlowFields ~\cite{Bailer2016FlowEstimation}, CPM - CPM Flow ~\cite{HuYinlinandSongRuiandLi2016}, DM - DeepMatching ~\cite{Weinzaepfel2013DeepFlow:Matching}, DF - DiscreteFlow ~\cite{Menze2015DiscreteFlow}}
\label{tab:finetuning}
\end{table}
Our network is trained in two phases. First, it is pre-trained on the flying chairs dataset using the FlowFields matching algorithm followed by fine tuning to the specific dataset and matching algorithm at hand. Table \ref{tab:finetuning} shows the performance of the networks trained only on the flying chairs dataset compared to the networks fine tuned on the Sintel training set with either the FlowFields matching algorithm or the same matching algorithms used for evaluation. The network performance is quite good even without fine tuning. However the fine tuning phase still improves the performance by a considerable margin. Fine tuning on FlowFields applied to Sintel yields results comparable to fine tuning on the evaluation algorithm. Finally, using a different matching algorithm for pre-training (DeepMatching, last line in table \ref{tab:finetuning}) does not improve the results. 
We, therefore, suggest the best practice for incorporating new matching algorithms with our method as follows: For most cases using the network trained and fine tuned on FlowFields as an out-of-the-box solution should be sufficient. For improved results, we suggest fine tuning on the specific dataset and matching algorithm. Pre-training on the specific matching algorithm applied to the flying chairs dataset is not necessary, although it could be beneficial in some cases.

\subsection{Benchmarks results}
\label{exp_benchmark}
\begin{table}[t]
% \begin{center}
\centering
\begin{tabular}{|l|c|c|c|c|}
\hline
Method & EPE & EPE-noc & EPE-occ  \\
\hline
\textbf{FF~\cite{Bailer2016FlowEstimation}+Ours} & \textbf{5.535} & 2.372 & 31.296\\
PGM-C (anon.) &  5.591 &  2.672& 29.389\\
RicFlow (anon.) &  5.620& 2.765 & 28.907\\
\textbf{CPM~\cite{HuYinlinandSongRuiandLi2016}+Ours} &  \textbf{5.627} &  2.594&  30.344 \\
FF$^{+}$~\cite{Bailer2016FlowEstimation} & 5.707 & 2.684& 30.356 \\
\textbf{DM~\cite{Weinzaepfel2013DeepFlow:Matching}+Ours} & \textbf{5.711} & 2.650 & 30.642\\
Deep DF~\cite{Guney2016ACCV} & 5.728 & 2.623 &  31.042\\
FN2-ft-s(anon.) & 5.739 & 2.752	& 30.108\\
SBFlow (anon.) & 5.734 & 2.676	& 30.654\\
FF~\cite{Bailer2016FlowEstimation}+Epic & 5.810 & 2.621& 31.799\\
SPM-BPv2 ~\cite{Li2016SPM-BP:MRFs} & 5.812 & 2.754 & 30.743\\
FullFlow~\cite{Chen2016FullGrids} & 5.895 & 2.838& 30.793 \\
CPM+Epic~\cite{HuYinlinandSongRuiandLi2016} & 5.960 & 2.990& 35.14 \\
FN2(anon.) & 6.016 & 2.977	& 30.807\\
GPC~\cite{Wang2016TheCollider} & 6.040 & 2.938& 31.309\\
% DiscreteFlow+Ours & 5.899 & 2.785 &31.294& 0.964 & 3.530 & 36.121 \\
\textbf{DF~\cite{Menze2015DiscreteFlow}+Ours} & \textbf{6.044} & 2.788 &32.581\\
DF~\cite{Menze2015DiscreteFlow}+Epic & 6.077 & 2.937 & 31.685 \\
DM+Epic~\cite{Revaud2015EpicFlow:Flow} & 6.285 & 3.060& 32.564 \\
%MR-FLOW (anon.) & 6.18 & 1.27 & 3.92 & 35.97 \\
%FGI~\cite{li2016fast} & 6.61 & 1.15 & 3.99 & 39.98 \\
%TF+OFM~\cite{TF+OFM} & TBD &  &  & \\
%Deep+R~\cite{deep+r} & 6.77 & 1.16 & 3.84 & 41.69\\
%PatchBatch CENT+SD + Ours & ??? & ??? & ??? & ??? \\
%PatchBatch CENT+SD + Epic ~\cite{patchbatch} & 6.78 & 0.72 & 3.06 & 45.86 \\
%DeepFlow2~\cite{revaud2015deepmatching} & 6.93 & 1.18 & 3.86 & 42.85 \\
\hline
\end{tabular}
% \end{center}
\caption{Leading results for the Sintel benchmark using the 'final' rendering pass. EPE-noc and EPE-occ are the EPE in non-occluded and occluded pixels respectively.}
\label{tab:mpi_results}
\end{table}

\begin{figure*}[t]
%\begin{center}
\centering
\includegraphics[width=1\linewidth]{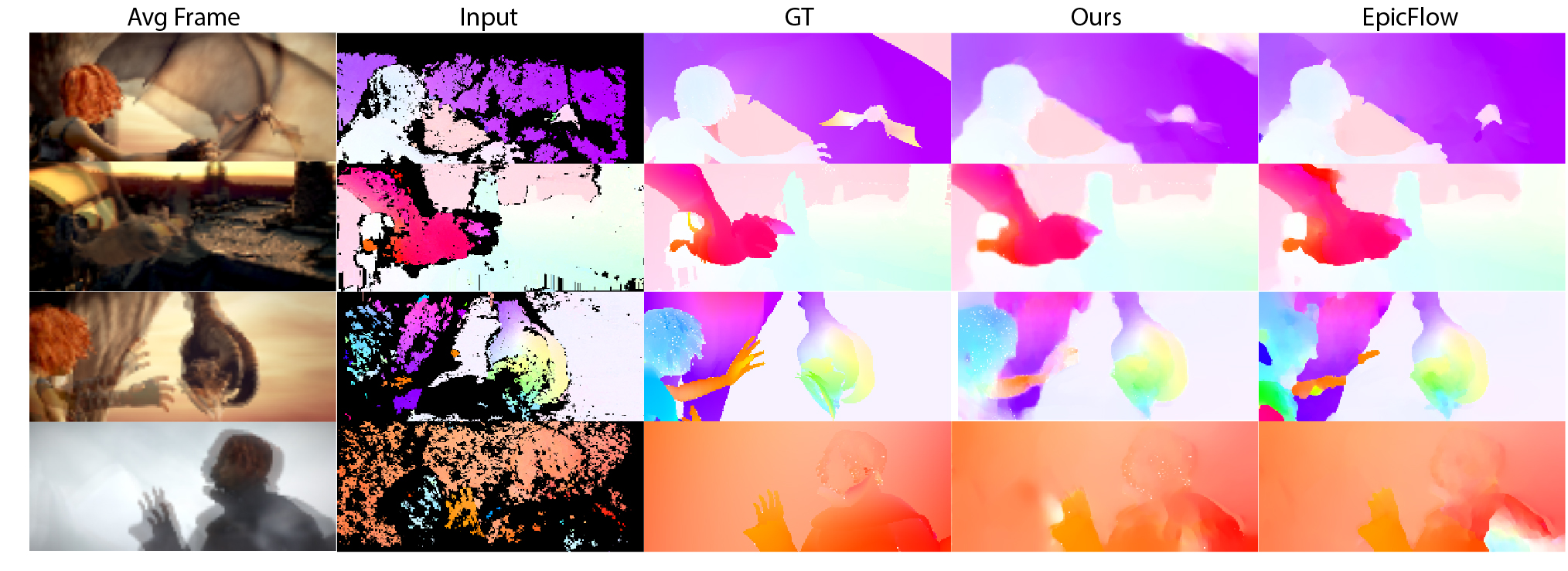}
%\end{center}
   \caption{A comparison of the predictions of our network to EpicFlow.} 
\label{fig:epic}
\end{figure*}
We applied our method to the output of several of the leading matching algorithms for Sintel and KITTI. The chosen matching algorithms are the highest on the leaderboards that have an available code and a reasonable running time. We used FlowField ~\cite{Bailer2016FlowEstimation}, DiscreteFlow ~\cite{Menze2015DiscreteFlow} and CPM-Flow ~\cite{HuYinlinandSongRuiandLi2016}. We also used DeepMatching, since it was used in the original EpicFlow paper ~\cite{Revaud2015EpicFlow:Flow}. 

For Sintel (Table \ref{tab:mpi_results}), we achieve state of the art results using FlowFields as the matching algorithm. For all the matching algorithms used, we achieve better results compared to EpicFlow improving the EPE by an average of 0.3px. Our performance is better in most areas including occluded, non-occluded and pixels in different distances from occlusion boundaries (with the exception of occluded pixels in CPM-flow and discrete flow). Figure ~\ref{fig:epic} shows a comparison of EpicFlow's and our outputs on several sparse flow maps produced by FlowFields for the Sintel validation set. Notice the performance difference in missing areas with noise (top right corner in the first row, the hand in the third and bottom right in the last row). Due to its non-learning nature, EpicFlow is clinging to any information that it finds within a segmented area and is, therefore, prone to fail in such regions. The flexibility of a data-driven algorithm, like ours, is more suitable here. Further analysis demonstrated the  superior performance of our method over EpicFlow in different regions (see supplementary).
% To further investigate our performance compared to EpicFlow, we looked at the EPE over all noisy pixels (pixels with EPE>3), occluded pixels and missing pixels (that did not appear in the input matches but are not occluded) from all the flow maps in the Sintel validation set. To make a fair comparison for this analysis, we performed our prediction without bidirectional averaging so the number of noisy and missing pixels in the input to our network and EpicFlow was identical. We found that our performance were better than EpicFlow in all of these areas, but it was significantly better only for the occluded pixels ($mean\pm sem$ difference between our Epic EPE and Our EPE: $0.08\pm0.1$, $0.44\pm0.33$, $1.42\pm0.61$ pixels; paired t-test p=0.43, p=0.19, p<0.05 for noisy, missing and occluded pixels respectively, n=167). 
Based on our results, we believe that applying our method to a matching algorithm ranked higher than FlowFields (ranked 7 before our contribution) should yield even better results.

\begin{table}[t]
\centering
\begin{small}
%\begin{center}
\begin{tabular}{|l|c|c|c|c|c|}
\hline
% &&&&\\
\multicolumn{1}{|c|}{Method} & \multicolumn{2}{|c|}{2012 - EPE} &\multicolumn{2}{|c|}{2012 - \%Out}& \multicolumn{1}{|c|}{2015} \\\cline{2-6}
  & \multicolumn{1}{|c|}{Noc} &   \multicolumn{1}{|c|}{All}&  \multicolumn{1}{|c|}{Noc} &   \multicolumn{1}{|c|}{All} &   \multicolumn{1}{|c|}{\%Out-All} \\
% Method & EPE & Out-occ\\
\hline
DF~\cite{Menze2015DiscreteFlow}+Ours & \textbf{1.0}& \textbf{2.4}& \textbf{4.94}&\textbf{14.13} & 23.55\\
DF+Epic & 1.3& 3.6& 6.23&16.63 &\textbf{22.38}\\
\hline
CPM+Ours & \textbf{1.0}& \textbf{2.5}&\textbf{5.28} & 14.57 & 23.84\\
CPM~\cite{HuYinlinandSongRuiandLi2016}+Epic & 1.3& 3.2& 5.79&\textbf{13.70} & \textbf{23.23}\\
\hline
FF+Ours & \textbf{1.1}&\textbf{2.6} & \textbf{5.57}& 14.76 &--\\
FF~\cite{Bailer2016FlowEstimation}+Epic &1.4 &3.5 &5.77 &\textbf{14.01}&--\\
\hline
DM+Ours & \textbf{1.1}&\textbf{2.7}& \textbf{5.85}& \textbf{15.03}& \textbf{24.65}\\
DM+Epic~\cite{Revaud2015EpicFlow:Flow} & 1.5& 3.8& 7.88& 17.08& 27.10\\
\hline
\end{tabular}
%\end{center}
\end{small}
\caption{KITTI 2012 and KITTI 2015 benchmarks results. The \%Out is the percentage of outlier pixels as defined by the benchmarks. FF does not have results on KITTI2015.
}
\label{tab:kitti_results}
\end{table}

% \begin{table}[t]
% \begin{center}
% \begin{tabular}{|l|c|}
% \hline
% Method & Out-Noc\\
% \hline\hline
% {\bf Imp.\ PatchBatch+SPCI} &  \textbf{4.65}\% \\
% CNN-HPM~\cite{bailer2016cnn} & 4.89\% \\
% {\bf Imp.\ PatchBatch} &  4.92\% \\
% PatchBatch+PS71~\cite{patchbatch} &  5.29\% \\
% PatchBatch~\cite{patchbatch} &  5.44\%\\
% PH-Flow~\cite{phflow} & 5.76\%\\
% FlowFields~\cite{flowfields} & 5.77\%\\
% CPM-Flow~\cite{huefficient} & 5.79\% \\
% \hline
% \end{tabular}
% \end{center}
% \caption{Top 8 published KITTI2012 Pure Optical Flow methods as of the submission date. Imp.\ PatchBatch denotes the PB pipeline with the improvements described in Section~\ref{sec:pb}. Out-Noc is the percentage of pixels with euclidean error $>$ 3 pixels out of the non-occluded pixels.}
% \label{tab:kitti12_results}
% \end{table}

For KITTI 2012 \cite{kitti2012}, using DiscreteFlow \cite{Menze2015DiscreteFlow} as the baseline matching algorithm, we achieve state-of-the-art results out of the published, pure optical flow methods, excluding semantic segmentation methods. We have the best performance, both in terms of EPE and the percentage outlier for non-occluded pixels (Table \ref{tab:kitti_results}). Compared to EpicFlow, the EPE is improved by a margin (21\%--33\%), using all matching algorithms. The \%Out measurement used in the KITTI datasets, calculates the percentage of pixels with $EPE > 3$. It is not linearly correlated with the EPE which we use as the target measurement, as reflected from the network's loss function. Consequently, while this measurement was improved for non-occluded pixels (3\%--25\%) we achieved mixed results for all pixels (-6\%--+15\%; Table \ref{tab:kitti_results}). Our results for KITTI 2015 \cite{Menze2015ObjectVehicles}, which uses only the \%Out as the evaluation system, were also mixed (Table \ref{tab:kitti_results}). The EPE measurement is not available in this benchmark, but our KITTI 2012 results support the possibility of an improvement in the EPE that is not reflected in the \%Out. The results for our validation set were better than EpicFlow using all the matching algorithms (see supplementary). 

%We evaluate our method only against methods not using additional information for the flow estimation, including those methods which used semantic segmentation.
\FloatBarrier
\subsection{Runtime analysis}

% \begin{table*}[!h]
% \begin{center}
% \begin{tabular}{|l|c|c|c|c|c|c|c|}
% \hline
% Step 		& Downsampling & Bidirectional & Edges 		& Network	& Upsampling & Variational 		& Total\\
% 	 		&				& average      &detection	& inference & 			 & post processing &\\
% \hline
% runtime (sec) & 0.058 		& 0.091 		& 0.150	& 0.025		& 0.009			 & 1  				& 1.334\\
% \hline
% \end{tabular}
% \end{center}
% \caption{Runtime of the different components of our solution measured for am image pair in the Sintel dataset}
% \label{tab:runtime}
% \end{table*}

\begin{table}[t]
%  \begin{minipage}{0.7\linewidth}
%    \begin{center}
\centering
	\begin{minipage}{0.7\linewidth}
      \begin{tabular}{|l|c|}
        \hline
        Step & runtime (sec)\\
        \hline
        Downsampling& 0.058\\
        Bi-directional average& 0.091\\
        Edges detection& 0.150\\
        Network inference& 0.025\\
        Upsampling& 0.009\\
        Variational post proc.& 1\\
        \hline
        Total & 1.333\\
        \hline
      \end{tabular}
     \end{minipage}
  %  \end{center}
%  \end{minipage}
%  \begin{minipage}{0.29\linewidth}
%  \end{minipage}
	\begin{minipage}{0.24\linewidth}
    \caption{Runtime of various steps of our solution for an image pair in the Sintel dataset.}
    \label{tab:runtime}
    \end{minipage}
\end{table}
Table \ref{tab:runtime} shows the runtime of the different components of our algorithm computed for one Sintel image pair (1024x436 pixels). The network inference ran on one NVIDIA GTX Titan black GPU (6GB RAM) while the other steps were performed on a single 3.4GHz CPU core. The run time of the edges detection and variational post-processing is as reported in \cite{Revaud2015EpicFlow:Flow}. The entire runtime was 1.333 seconds. This is slightly better than the reported runtime for EpicFlow (1.4 seconds). Notably, several parts in the pipeline could be dropped for better runtime without a big decrease in performance. The bi-directional average can be dropped in inference time (which will also reduce the downsampling by half), as well as the variational post processing, leaving the edges detection as the biggest bottleneck. Therefore, without much performance loss, our method can be as fast as 5 fps. Combined with a fast edge detection and matching algorithm, future work can produce a real-time optical flow algorithm.          

\section{Conclusions}
Using a fully convolutional neural network, we have presented a data-driven solution for sparse-to-dense interpolation for optical flow producing state-of-the-art results. Our solution was inspired by ideas taken from interpolation processes in the visual cortex. We embedded anatomical features, like lateral dependency and multi-layer processing, by using the loss function, thereby applying supervision rather than using the architecture of the network which contributes to the simplicity of our solution. We also showed that the edge information is crucial for learning to interpolate. The network learns to use the edges as stoppers for the spread of interpolation, much like in the visual cortex.

Our solution is robust and can be applied to the output of different  matching algorithms and our code and models will be made completely public. We encourage new solutions to use our method as part of their pipeline. 

\section{Acknowledgments}
This research is supported by the Intel Collaborative Research Institute for Computational Intelligence (ICRI-CI) and by the Israeli Ministry of Science, Technology, and Space.
% \begin{figure}
% \centering
%     \begin{subfigure}{.4\textwidth}
%         \centering
%         \includegraphics[width=\linewidth]{Edges1.jpg}
%         \caption{}\label{fig:fig_a}
%     \end{subfigure} %
%     \begin{subfigure}{.4\textwidth}
%         \centering
%         \includegraphics[width=\linewidth]{Edges2.jpg}
%         \caption{}\label{fig:fig_b}
%     \end{subfigure} %
%     \begin{subfigure}{\textwidth}
%         \centering
%         \includegraphics[width=\linewidth]{Edges1.jpg}
%         \caption{}\label{fig:fig_c}
%     \end{subfigure}
% \caption{Some general caption of all the figures. In (\subref{fig:fig_a}) you can see a green square....}
% \end{figure}
%Drawing further ideas from the visual cortex we note that the edges are calculated in parallel within the process, and not as received as an outside input. Additionally, in some cases like some illusions the edges are not part of the input at all and are inferred by a top down information due to good continuity and other concepts. Future work could find a way to incorporate the edges extraction from the original image as part of the network, as well as teaching the network to complete missing boundaries.
%Finally, As we are only part of the optical flow pipeline, combining a deep learning based matching algorithm with our interpolation network, would form a fast and accurate end-to-end solution for optical flow estimation.   
\FloatBarrier
{\small
\bibliographystyle{ieee}
\bibliography{Mendeley}
}
\clearpage
\appendix
\title{InterpoNet, A brain inspired neural network for optical flow dense interpolation - Supplementary}
\author{}
\date{}
\maketitle
\section{Bi-directional Averaging}
\renewcommand\thetable{S.\arabic{table}}    
\setcounter{table}{0}
\begin{table}[h]
\begin{center}
\begin{tabular}{|l|c|c|c|c|}
\hline
% &&&&\\
& FF & DF & CPM & DM\\ 
\hline
Ours bidi & 5.470& 6.142& 5.851&5.971\\
Ours no-bidi &  5.363& 6.141& 5.768&6.017\\
Epic bidi & 6.225& 6.837& 6.521&6.261\\
Epic no-bidi & 5.815 & 6.625& 6.337 & 6.441\\
\hline
\end{tabular}
\end{center}
\caption{Comparison of the results of our method and EpicFlow for the Sintel validation set with and without applying bi-directional averaging to the input in evaluation time .}
\label{tab:bidi_avg}
\end{table}

We found that the training process results declined when the average number of missing pixels in the training flow maps was too high. Some of the matching algorithms, in particular DeepMatching, did produce sparse maps like these. To tackle this problem, we calculate the flow map bi-directionally (From $I$ to $I'$ and from $I'$ to $I$) using the matching algorithm. We invert the second flow map and average the two maps. This simple step solves the sparseness problem for all of the matching algorithms we used. This procedure added to the computation time of our method. However most matching algorithms already compute bi-directional maps for consistency check and false matches filtering purposes and so we did not need to apply them twice. Importantly, we found that the bi-directional averaging is critical mostly for training the network and specifically for DeepMatching outputs. Training on FlowFields non averaged maps, for instance, gives comparable results to training with the averaged maps. Interestingly, applying EpicFlow on the bi-directional average of the DeepMatching algorithm output also slightly improved their results (Table \ref{tab:bidi_avg}). For consistency reasons, we choose to present in this paper the results gained using the bi-directional averaged maps for training and evaluation. However, for all matching algorithms using only the original, non averaged map, in evaluation time yields results similar to those presented (Table \ref{tab:bidi_avg}). The analysis in this section was performed without the variational post processing for both our method and EpicFlow.

\section{Choice of Training and Validation Sets}
The validation sets for both KITTI2012 and KITTI2015 datasets were the last 20\% of the pairs in each. For the Sintel dataset, due to the temporal dependencies within scenes which are a pitfall for over-fitting, we define 4 whole scenes including 167 image pairs as a validation set rather than a random sample. We use the same validation set in the pre-training and Sintel fine tuning phases. 

\section{Early Stopping}
Early stopping served as our only regularization method. The number of steps before performing the stop was 5000,1000 and 400 for training on the flying chairs, Sintel and KITTI datasets respectively. We use 4 rounds of early stopping in which we divide the learning rate by two starting with a learning rate of $5\times 10^{-5}$ for the pre-training and $5\times 10^{-6}$ for the fine tuning. After 4 rounds, we choose the weights that yielded the best performance on the validation set throughout the training. 

\section{Quantitative comparison To EpicFlow}
\begin{table}[h]
\begin{center}
\begin{tabular}{|l|c|c|c|c|}
\hline
% &&&&\\
\multicolumn{1}{|c|}{Method} & \multicolumn{2}{|c|}{KITTI 2012}  &  \multicolumn{2}{|c|}{KITTI 2015}\\\cline{2-5}
  & \multicolumn{1}{|c|}{EPE} &   \multicolumn{1}{|c|}{\%Out-all}&  \multicolumn{1}{|c|}{EPE} &   \multicolumn{1}{|c|}{\%Out-all} \\
% Method & EPE & Out-occ\\
\hline
% PatchBatch + Ours \footnote{\label{note_pb}The validation set for the PatchBatch algorithm consisted of only half of the validation set for the other algorithms}&2.287&9.00 & 9.552&27.75\\
% PatchBatch~\cite{Gadot2016PatchBatch:Flow}+Epic & 2.745&10.09 & 13.378&29.20\\
% \hline
FF+Ours & \textbf{2.363}&\textbf{11.11} & \textbf{7.921}& \textbf{29.00}\\
FF+Epic &3.518 &11.25 &16.100 &33.00\\
\hline
CPM+Ours & \textbf{2.271}&11.3 &\textbf{6.92} & \textbf{26.04}\\
CPM+Epic & 3.337& \textbf{11.16}&15.135 &32.48\\
\hline
DF+Ours & \textbf{2.074}& \textbf{9.01}& \textbf{6.626}&\textbf{24.29}\\
DF+Epic & 2.92&12.34 & 11.680&30.34\\
\hline
DM+Ours & \textbf{2.168}&\textbf{9.57}& \textbf{6.733}& \textbf{28.84}\\
DM+Epic &3.515 &14.20 &14.068 &35.12\\
\hline
\end{tabular}
\end{center}
\caption{Comparison of our model to EpicFlow on the KITTI 2012 and KITTI 2015 validation sets. The \%Out is the percentage of pixels with $EPE>3$ pixels.}
\label{tab:kitti_val_results}
\end{table}

Table \ref{tab:kitti_val_results} shows the results gained using our method compared to EpicFlow for both KITTI datasets. Our method surpassed EpicFlow in all measurements  (excluding \%Out for KITTI 2012 using CPM).
To further investigate our performance compared to EpicFlow, we looked at the EPE over all noisy pixels (pixels with $EPE>3$) and missing pixels from all the flow maps in the Sintel validation set. To make a fair comparison for this analysis, we performed our prediction without bi-directional averaging so the number of noisy and missing pixels in the input to our network and EpicFlow was identical. We found that our performance were better than EpicFlow's in both of these areas, but it was significantly better only for the missing pixels ($Mean\pm SEM$ difference between Epic EPE and Our EPE: $0.08\pm0.1$, $1.11\pm0.42$ pixels; paired t-test p=0.42, $p<0.01$ for noisy and missing pixels respectively, n=167). This emphasize our superiority over EpicFlow, Especially in large missing regions, as was demonstrated in Figure 5 of the main text.

\section{Supplemental Figures}
The supplemental figures presented here show further examples on top of the ones presented in the figures in the main text. Figure \ref{fig:layers_supp} shows the progression of the prediction process in the network as appears in the output of the different detour layers, similar to figure 3a in the main text. Notice here also how the network first performs a simple interpolation and then refines the predictions in the deeper layers. Figure \ref{fig:edges_supp} presents the predictions of networks with and without the edges input, similar to Figure 4a in the main text. The progression of the predictions in the different layers in those network is presented in figure \ref{fig:edges_layerssupp}. These two figures illustrate how the edges input function in the network - acting as a stopper for spread of activation. Notice how the bottom "simple interpolation" layers perform similarly in both networks. However, starting from layer 4, the refinement process is very different. The network that receives the edges as input utilizes them to act as motion boundaries. Finally, figures \ref{fig:epic_supp}, \ref{fig:epic_kitti2012_supp} and \ref{fig:epic_kitti2015_supp} shows additional examples to the ones presented in figure 5 in the main text, for the comparison between the performance of our method and EpicFlow on the Sintel, KITTI 2012 and KITTI 2015 validation sets. 
\renewcommand\thefigure{S.\arabic{figure}}    
\clearpage
\setcounter{figure}{0}
\begin{figure*}
\centering
	\begin{subfigure}{0.75\linewidth}
        \centering
        \includegraphics[width=\linewidth]{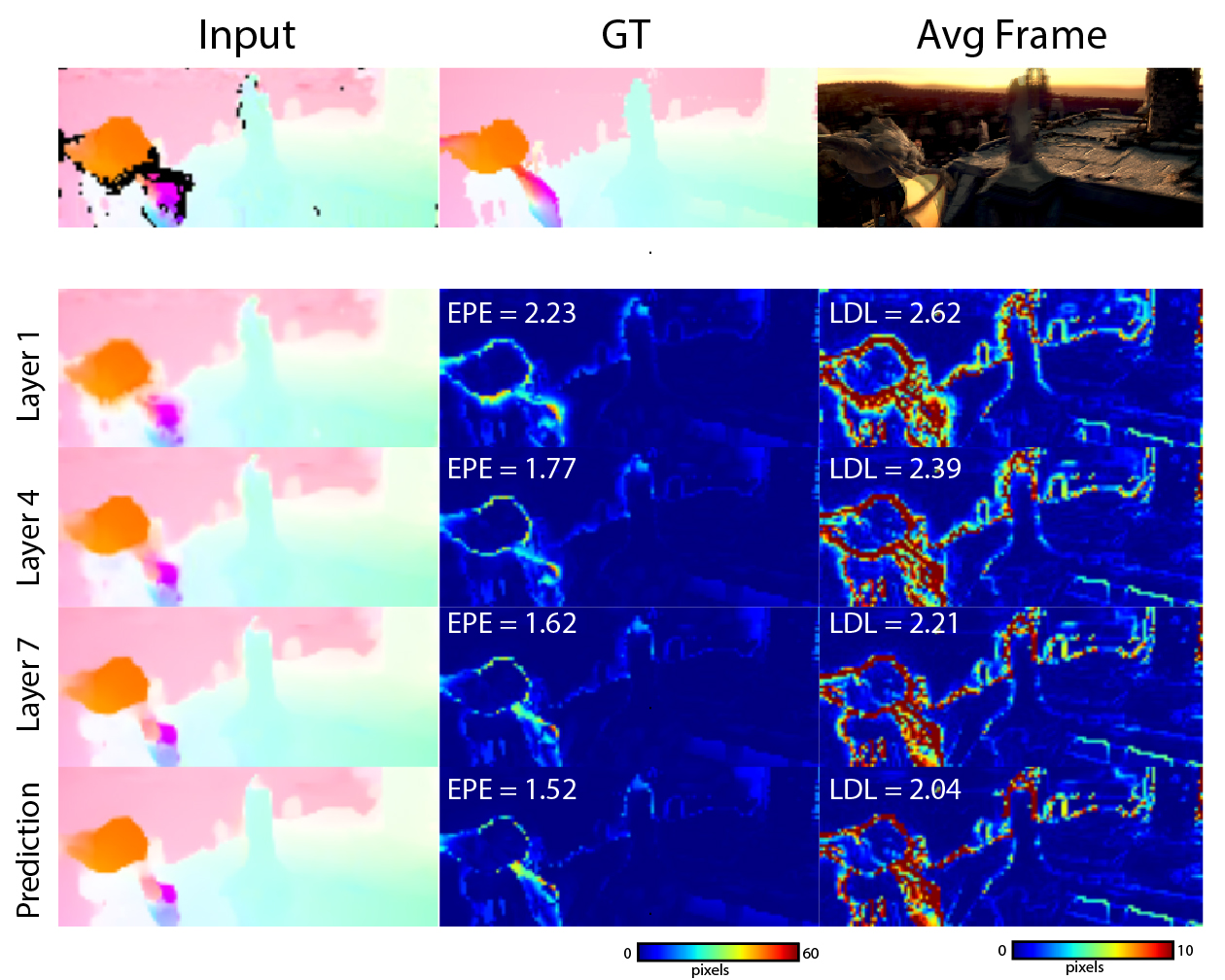}
    \end{subfigure} 
    \begin{subfigure}{0.75\linewidth}
        \centering
        \includegraphics[width=\linewidth]{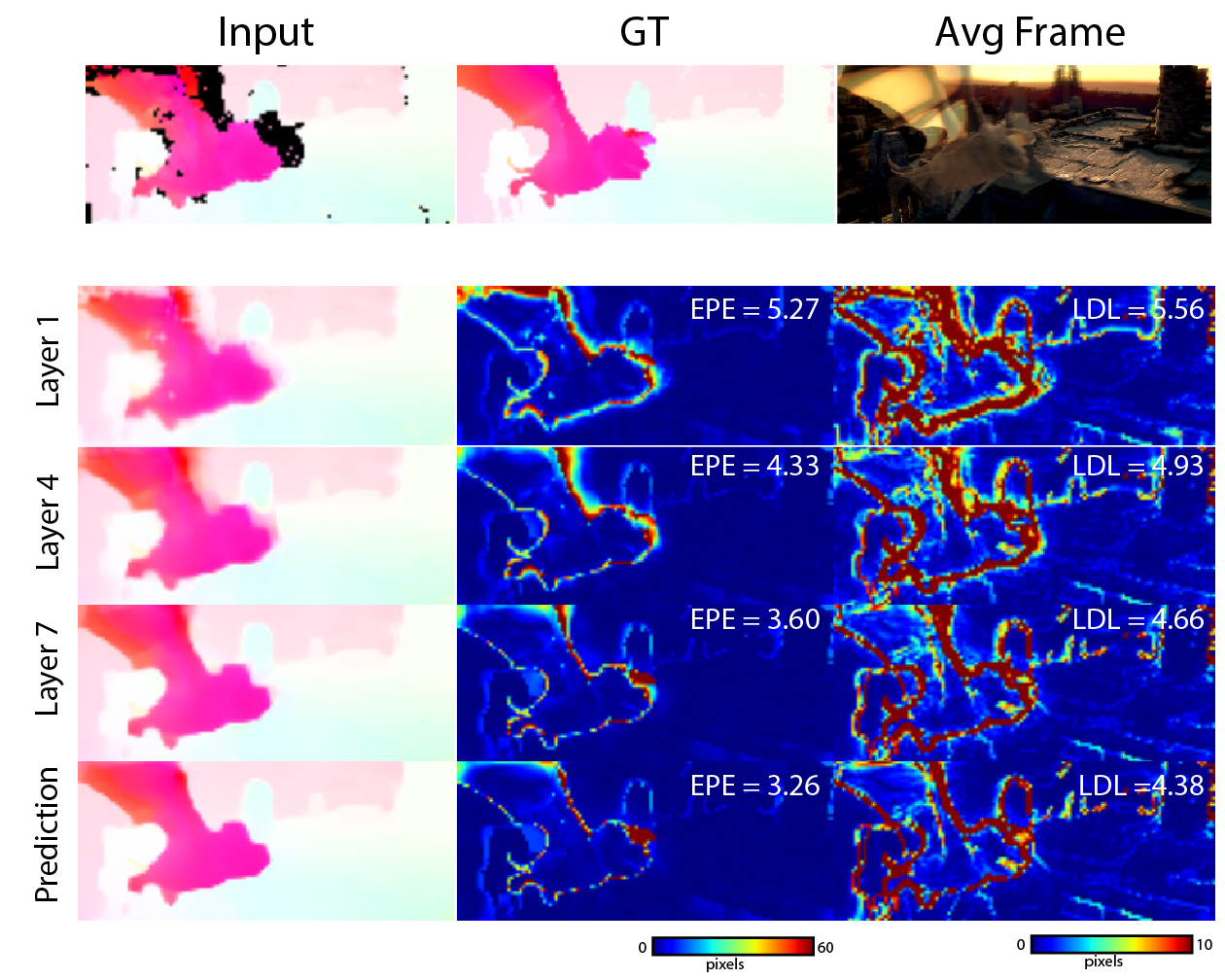}
    \end{subfigure} %
	\caption{Predictions in different layers -- additional examples to figure 3a in the main text. The progression of the prediction process throughout the different layers in the network, as shown by the detour networks outputs. Starting from the second row, the second and third columns are the EPE and LD loss maps respectively.}\label{fig:layers_supp}
\end{figure*}

\begin{figure*}
\centering
	  \begin{subfigure}[t]{0.85\linewidth}
          \centering
          \includegraphics[width=\linewidth]{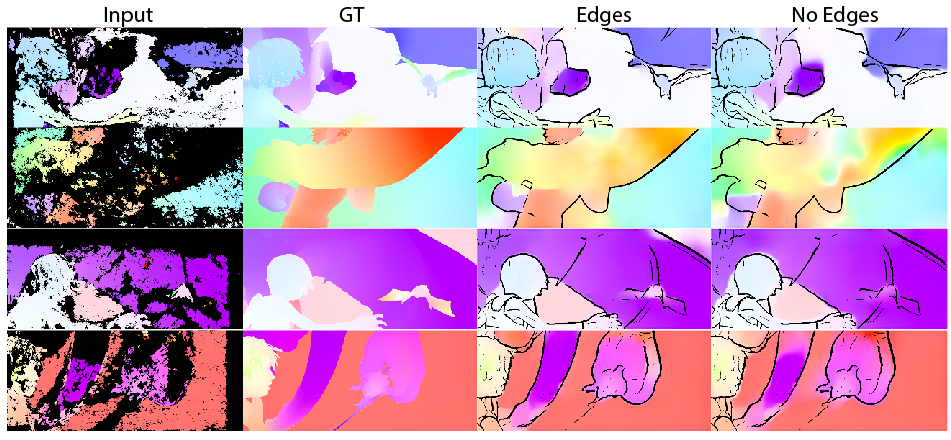}
      \end{subfigure}  
	
    \caption{The predictions of the networks with and without the edges input (additional examples to figure 4a in the main text). Edges are marked with black lines. }
	\label{fig:edges_supp}
\end{figure*}

\begin{figure*}
\centering
	\begin{subfigure}{0.40\linewidth}
        \centering
        \includegraphics[width=\linewidth]{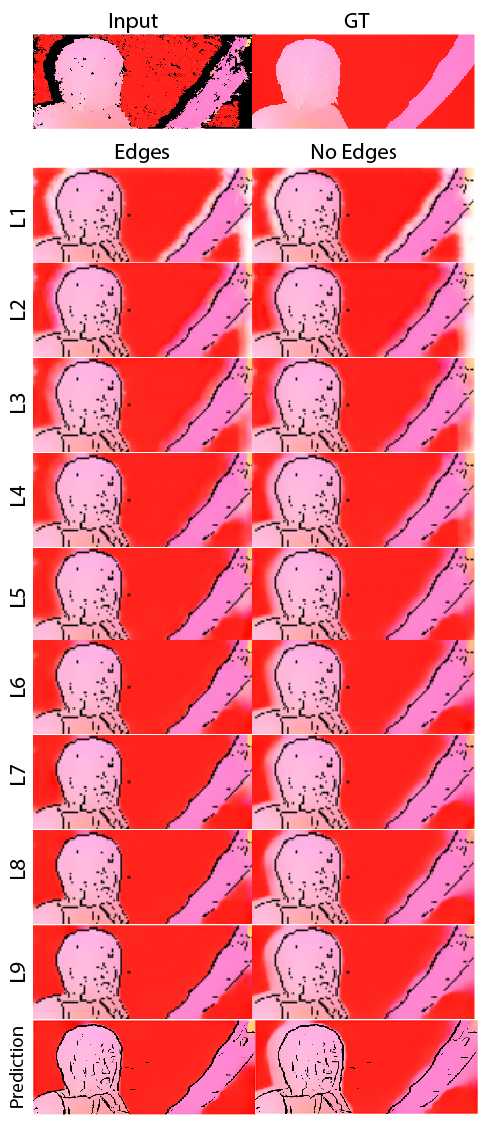}
    \end{subfigure} %
    \caption{The progression of the prediction process throughout the different layers in the network, as shown by the detour networks outputs for networks with and without the edges input, notice the similarity in the bottom layers and then the divergence starting from layer 4.}
	\label{fig:edges_layerssupp}
\end{figure*}

\begin{figure*}[t]
% \begin{center}
\centering
\includegraphics[width=1\linewidth]{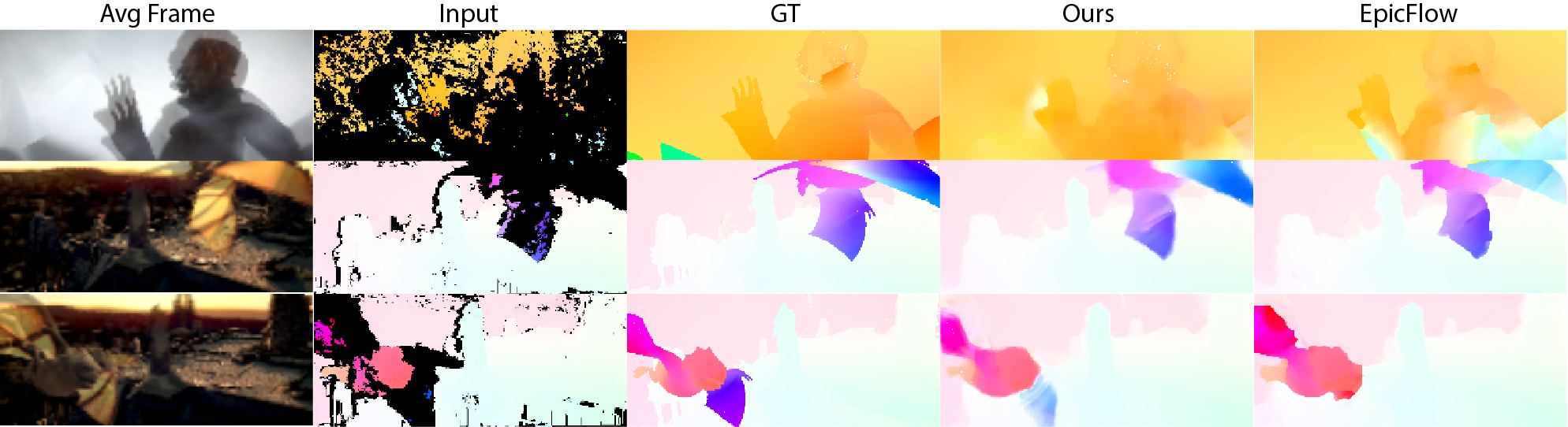}
% \end{center}
   \caption{A comparison of the predictions of our network to EpicFlow on examples from the Sintel validation set. (additional examples to figure 5 in the main text).} 
\label{fig:epic_supp}
\end{figure*}

\begin{figure*}
\centering
% \begin{center}
\includegraphics[width=1\linewidth]{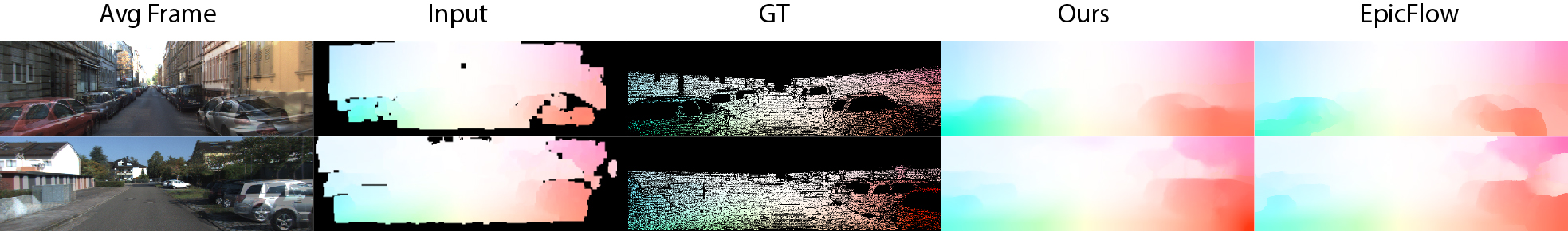}
% \end{center}
   \caption{A comparison of the predictions of our network to EpicFlow on examples from the KITTI 2012 validation set.} 
\label{fig:epic_kitti2012_supp}
\end{figure*}

\begin{figure*}
\centering
% \begin{center}
\includegraphics[width=1\linewidth]{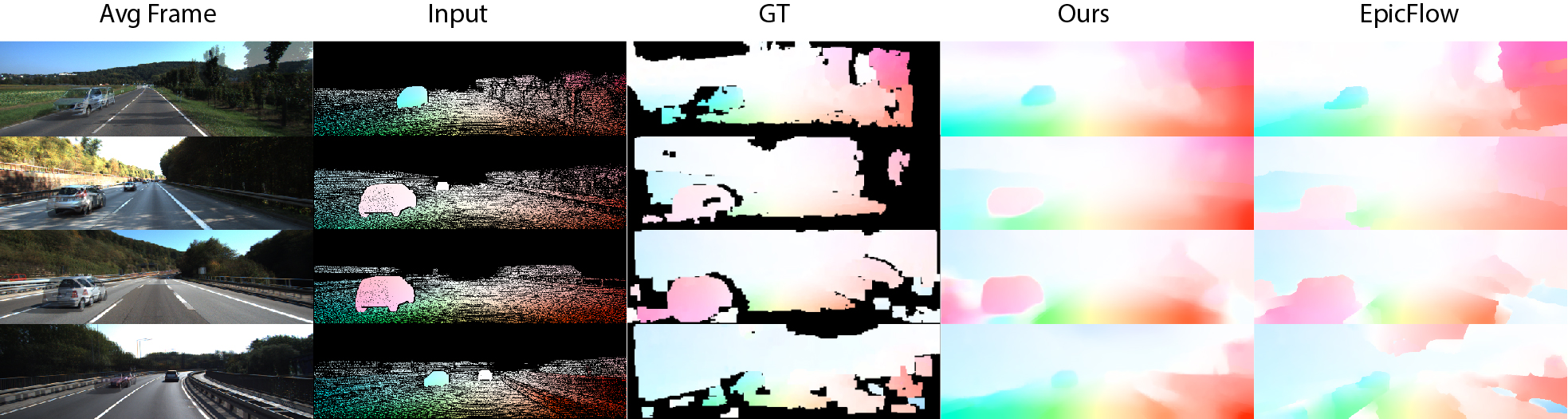}
% \end{center}
   \caption{A comparison of the predictions of our network to EpicFlow on examples from the KITTI 2015 validation set.} 
\label{fig:epic_kitti2015_supp}
\end{figure*}

\end{document}